\documentclass[conference]{IEEEtran}
\IEEEoverridecommandlockouts
\usepackage{cite}
\usepackage{amsmath,amssymb,amsfonts}
\usepackage{algorithmic}
\usepackage{algorithm}
\usepackage{xcolor}
\usepackage{graphicx}
\usepackage{textcomp}
\usepackage{url}
\usepackage{hyperref}
\newtheorem{prop}{Proposition}

\newtheorem{theorem}{Theorem}
\newtheorem{definition}{Definition}
\usepackage{subcaption}
\usepackage{caption}
\usepackage{booktabs}
\usepackage{multirow}
\makeatletter
\def\BibTeX{{\rm B\kern-.05em{\sc i\kern-.025em b}\kern-.08em
    T\kern-.1667em\lower.7ex\hbox{E}\kern-.125emX}}
\begin{document}

\title{Obtaining Dyadic Fairness by Optimal Transport\\
\thanks{This work was supported in part by STCSM (No. 20DZ1100300) and NSFC (No. 12071145). This paper is funded by Shanghai Trusted Industry Internet Software Collaborative Innovation Center. Corresponding author: Xiangfeng Wang.}
}


\author{\IEEEauthorblockN{Moyi YANG\IEEEauthorrefmark{1},
Junjie SHENG\IEEEauthorrefmark{1},
Wenyan LIU\IEEEauthorrefmark{1}, 
Bo JIN\IEEEauthorrefmark{1}, 
Xiaoling WANG\IEEEauthorrefmark{1},
Xiangfeng WANG\IEEEauthorrefmark{1}}
\IEEEauthorblockA{\IEEEauthorrefmark{1}School of Computer Science and Technology, East China Normal University, Shanghai, China 200062}
}


\maketitle

\begin{abstract}
Fairness has been taken as a critical metric in machine learning models, which is considered as an important component of trustworthy machine learning.
In this paper, we focus on obtaining fairness for popular link prediction tasks, which are measured by {\em{dyadic fairness}}.
A novel pre-processing methodology is proposed to establish dyadic fairness through data repairing based on optimal transport theory.
With the well-established theoretical connection between the dyadic fairness for graph link prediction and a conditional distribution alignment problem, the dyadic repairing scheme can be equivalently transformed into a conditional distribution alignment problem.
Furthermore, an optimal transport-based dyadic fairness algorithm called \textbf{DyadicOT} is obtained by efficiently solving the alignment problem, satisfying flexibility and unambiguity requirements.
The proposed \textbf{DyadicOT} algorithm shows superior results in obtaining fairness compared to other fairness methods on two benchmark graph datasets.
\end{abstract}


\section{Introduction}
Machine learning has been widely adopted in real-world applications.
Although remarkable results were achieved in the prediction and decision-making scenarios, unexpected bias occurs regularly \cite{stoica2018algorithmic,besse2018confidence,friedler2019comparative}.
For example, the famous new media company ProPublica found that black defendants were far more likely than white defendants to be incorrectly judged as having a higher risk of recidivism in the COMPAS system\cite{angwin2016machine}.
The Amazon company found that the AI hiring tool they developed to automate the hiring process is biased against women\cite{lauret2019amazon}.
Many works emerge to design algorithms to avoid such biases and aim to obtain {\em{fair}} machine learning models.

This work focuses on achieving fairness in link prediction tasks.
The link prediction task is a fundamental but essential problem in modern machine learning applications, not limited to recommendation systems and knowledge graph completion.
The main goal is to predict whether the link between two nodes exists in a graph.
Many existing popular algorithms, e.g., Node2Vec\cite{grover2016node2vec} and GCN\cite{kipf2016semi}, have been proposed to solve the link prediction task with superior performance in many scenarios.
However, the dataset collected for the model training procedure usually has various unexpected biases. This will lead to unfair results for the link prediction model obtained.
For instance, after collecting data from social media platforms, early works highlighted that users were more interested in conversing with others of the same race and gender\cite{khanam2020homophily}.
Link prediction models, trained based on such unfair data, will also tend to predict the existence of links between nodes with the same sensitive information.
This will unfairly disadvantage some users.
To formally define such an unfair phenomenon, \cite{li2021on,DBLP:conf/aaai/MasrourWYTE20} introduced dyadic fairness for link prediction of graphs.
The dyadic fairness criterion expects the prediction results to be independent of the sensitive attributes from the given two nodes.

Recently, several works have been proposed to achieve dyadic fairness in link prediction tasks, which can be roughly divided into three categories:
1) in-processing scheme\cite{li2021on} considers modifying the learning algorithm to eliminate bias;
2) post-processing scheme\cite{DBLP:conf/aaai/MasrourWYTE20} attempts to debias directly the model's output after training;
3) pre-processing scheme\cite{DBLP:journals/corr/abs-2104-14210} aims to repair the graph data before the training procedure, and ensures the link prediction results can satisfy dyadic fairness.
In this paper, our proposed method is established under the pre-processing scheme.
Compared to the in-processing and post-processing schemes, the pre-processing scheme should be the most flexible fairness intervention\cite{nielsen2020practical}.
Suppose the discriminating information is removed from the data during the pre-processing stage, the processed data could be utilized to solve arbitrary downstream tasks without concern about the fairness issue.
Few works have studied obtaining dyadic fairness through a pre-processing scheme.
FairDrop\cite{DBLP:journals/corr/abs-2104-14210} proposed a heuristic repairing method that can mask out edges based on the dyadic sensitive attributes.
It is easy to implement but without a theoretical guarantee of achieving fairness.
To design a theoretically sound pre-processing scheme, FairEdge\cite{DBLP:journals/corr/abs-2010-16326} firstly adopts the Optimal Transport (OT) theory\cite{villani2009optimal} to justify whether dyadic fairness can be obtained through a repairing scheme.
FairEdge focuses on the plain graph (the node has no attribute) and proposes to repair adjacency information distributions (conditioned on sensitive attribute) to the corresponding Wasserstein barycenter.
Dyadic fairness is obtained once the adjacency information distributions are all repaired as the obtained Wasserstein barycenter.
Unlike the previous approach, we expect to focus on attributed graphs (each node has attributes) that are more general in the real world. 
Because node attributes introduce bias even if the bias of adjacency information can be removed, those algorithms that simply consider plain graphs cannot solve this problem, and the achievement of dyadic fairness on attributed graphs is still underexploited.

\section{Related Works}
\subsection{Fairness in Link Prediction}
Link prediction is a well-researched problem in applications related to graph data\cite{al2006link,masrour2015network}. Since fairness in graph-structured data is a relatively new research topic, only a few works have investigated fairness issues in link prediction.
In\cite{DBLP:journals/corr/abs-2104-14210}, the authors proposed a biased dropout strategy that forces the graph topology to reduce the homophily of sensitive attributes.
Meanwhile, to measure the improvements for the link prediction, they also defined a novel group-based fairness metric on dyadic level groups.
In contrast, \cite{DBLP:conf/aaai/MasrourWYTE20} considered generating more heterogeneous links to alleviate the filter bubble problem.
In addition, they further presented a novel framework that combines adversarial network representation learning with supervised link prediction.
Following the idea of adversarially removing unfair effects, \cite{li2021on} proposes the algorithm FairAdj to empirically learn a fair adjacency matrix with proper graph structural constraints for fair link prediction to ensure predictive accuracy as much as possible simultaneously.
Most similar to our method, \cite{DBLP:journals/corr/abs-2010-16326} formulated the problem of fair edge prediction and proposed an embedding-agnostic repairing procedure for the adjacency matrix with a trade-off between group and individual fairness.
However, they still ignore the node attributes, which impact both the prediction and fairness performance.

\subsection{Fairness with Optimal Transport}
In the context of ML fairness, several works have proposed using the capacity of optimal transport to align probability distributions, overcoming the limitation of most approaches that approximate fairness by imposing constraints on the lower-order moments.
Along with this motivation, most of the existing methods consider using optimal transport theory to match distributions corresponding to different sensitive attributes in the model input space or the model output space, which corresponds to pre-processing\cite{pmlr-v97-gordaliza19a, feldman2015certifying, DBLP:journals/corr/abs-2010-16326} and post-processing\cite{jiang2019wasserstein, chzhen2020fair} methods, respectively. 
In addition, the in-processing\cite{jiang2019wasserstein, chiappa2021fairness} methods based on optimal transport achieve fairness by imposing constraints in terms of the Wasserstein distance in the objective function.

\section{Dyadic Fairness in Link Prediction}
In this section, we formulate dyadic fairness in the link prediction task and define two metrics (dyadic disparate impact and dyadic balanced error rate) to quantify dyadic fairness.
Then we conclude two desired properties for our repairing algorithm that try to obtain dyadic fairness, i.e., flexibility and unambiguity.
We further theoretically discuss how these properties can be achieved and prove that aligning conditional attribute and adjacency distributions to the same distribution can obtain dyadic fairness with these properties.

\subsection{Problem Formulation}
Given the graph $\mathcal{G}:= \left( \mathcal{V}, \mathcal{E} \right)$ with $\mathcal{V}:= \{v_1, \dots, v_N\}$ be the node set of the graph and  $\mathcal{E}:= \{e_1, \dots, e_N\}$ be the edge set of the graph.
Each node $v_i$ be endowed with a vector $\mathbf{x}_i\in\mathbb{R}^M$ of attributes.
Each edge $e_i$ is the $i$th row of a non-negative adjacency matrix $A\in\{0,1\}^{N\times N}$ which summarizes the connectivity in the graph.
If nodes $v_i$ and $v_{j}$ are connected, then $A_{ij}=1$; otherwise, $A_{ij}=0$.
The link prediction model usually identifies whether the link between two nodes ($i,j$) exists based on their node representations, i.e.,  $g: \boldsymbol{z}_i \times \boldsymbol{z}_j \mapsto \{0, 1\}$ where the $\boldsymbol{z}_i$ denotes the node $i$'s representation.
The $\boldsymbol{z}_i$ is usually obtained by random walk or graph convolution on the whole graph: $\boldsymbol{z}_i = f(\mathcal{G})[i]$ where the $f: \mathcal{G} \mapsto \mathbb{R}^{N\times d}$ is called the embedding function.
The $d$ is the dimension of the node representation, and the $f$ can be Node2Vec, GCN, GAT, etc.
The link predictor $g$ takes two nodes' representations with the node representations and directly outputs whether a link exists between them.
To study the fairness of link prediction tasks, we assume that all nodes have one sensitive feature  $S: \mathcal{V} \rightarrow \mathcal{S}$.
We also take the binary sensitive feature $\mathcal{S}=\{0, 1\}$ first and let $S(i)$ denote the sensitive feature of node $i$.
The binary sensitive feature will be relaxed later.
Before proposing our algorithm, we make the following two assumptions:

\smallskip
\noindent{1). {\bf{Equivalence assumption}}} $$\mathbb{P}\left(S\oplus S^\prime=1 \right)=\mathbb{P} \left(S\oplus S^\prime=0\right)=\frac{1}{2},$$which is based on the fact that each node has the same chance of being sampled regardless of its sensitive attribute value. 
For instance, $\mathbb{P}(S=man)=\mathbb{P}(S=woman)$ is always an equivalence relationship independent of the sampling process and the obtained graph data itself;

\smallskip
\noindent{2). {\bf{Propensity assumption}}}
\begin{eqnarray}
&&\mathbb{P}\left(g(\boldsymbol{z}_u,\boldsymbol{z}_v) = 1\,\big| \,S(u)\oplus S(v)=0\right)\nonumber \\ & & \quad\quad \geq\mathbb{P} \left(g(\boldsymbol{z}_u,\boldsymbol{z}_v)=1 \,\big| \,S(u)\oplus S(v)=1 \right),\nonumber
\end{eqnarray}which illustrates that the classifier we consider here will tend to predict the existence of links between nodes with the same sensitive attributes.

For link prediction problems, the main unfairness phenomenon is assigning high link probability to nodes with the same sensitive feature while assigning low probability to nodes with different sensitive features.
For example, a user may be treated unfairly on social platforms because they are rarely recommended to users of a different gender or race.
This unfairness can be defined mathematically as in \cite{li2021on}.
\begin{definition}
[Dyadic Fairness] A link predictor $g$ obtains dyadic fairness if for node representation $\boldsymbol{z}_i$ and $\boldsymbol{z}_j$
\begin{equation}\label{eq: dp}
\mathbb{P}\left( g(\boldsymbol{z}_i, \boldsymbol{z}_j)\, \big| \, S(i)\oplus S(j)=1 \right) = \mathbb{P} \left( g(\boldsymbol{z}_i, \boldsymbol{z}_j) \, \big|\, S(i) \oplus S(j)=0 \right).
\end{equation}
\end{definition}When the link predictor decides the link between two nodes in the same proportion regardless of whether they have the same sensitive attributes, the predictor can be denoted as obtaining dyadic fairness.
Actually, the dyadic fairness described in \eqref{eq: dp} is difficult to achieve in real data. 
Therefore, to better quantify fairness, we could adopt two other essential fairness metrics, i.e., {\em{dyadic disparate impact}} (DDI) and {\em{dyadic balanced error rate}} (DBER), which are defined as follows:
\begin{definition}[DDI: Dyadic Disparate Impact]
Given a graph $\mathcal{G}=(\mathcal{V},\mathcal{E})$ and a function $g(\boldsymbol{z}_u,\boldsymbol{z}_v):\mathbb{R}^d\times\mathbb{R}^d\rightarrow\{0,1\}$, we define the link prediction function $g$ has Disparate Impact at level $\tau\in (0,1]$ on $S(u)\oplus S(v)$ w.r.t.$\boldsymbol{Z}$ if:
\begin{equation}
    \mathrm{DDI}\left(g,\boldsymbol{Z},\mathcal{S}\right)=\frac{\mathbb{P}\left( g(\boldsymbol{z}_u,\boldsymbol{z}_v)=1\,\big|\,S(u)\oplus S(v)=1 \right)}{\mathbb{P}\left( g(\boldsymbol{z}_u,\boldsymbol{z}_v)=1\,\big|\, S(u)\oplus S(v)=0 \right)}\leq\tau.
    \label{eq: di}
\end{equation}
\end{definition}
DDI measures the fairness level of the predictor. 
The higher the value of $\tau$, the fairer it is. 
Ideally, when the value of $\tau$ reaches $1$, it means that the link predictor achieves dyadic fairness.
\begin{definition}[DBER: Dyadic Balanced Error Rate]
For a graph $\mathcal{G}=(\mathcal{V},\mathcal{E})$ and a function $g(\boldsymbol{z}_u,\boldsymbol{z}_v):\mathbb{R}^d\times\mathbb{R}^d\rightarrow\{0,1\}$, we define the dyadic balanced error rate of the predictor $g$ as the average class-conditional error:
\begin{equation}
\begin{aligned}
& \mathrm{DBER}\left(g,\boldsymbol{Z},\mathcal{S} \right)= \frac{1}{2} \left[\mathbb{P} \left( g(\boldsymbol{z}_u,\boldsymbol{z}_v)=0\, \big|\, S(u)\oplus S(v)=1 \right)\right.\\
&\qquad  \left. + \mathbb{P}\left( g(\boldsymbol{z}_u,\boldsymbol{z}_v)=1\, \big|\, S(u)\oplus S(v)=0 \right)\right].
\end{aligned}
\label{eq:ber}
\end{equation}
\end{definition}
DBER measures the general misclassification error of sensitive attributes by $g$ in the particular case of $\mathbb{P}(S\oplus S^\prime=1)=\mathbb{P}(S\oplus S^\prime=0)=\frac{1}{2}$.
DBER can be guaranteed to be smaller than $\frac{1}{2}$.
With a larger DBER, the data and predictor $g$ will be more fair.
If DBER equals $\frac{1}{2}$, then DDI will be $1$, and dyadic fairness will be achieved.

\subsection{Obtaining Dyadic Fairness}
In this paper, we consider establishing dyadic fairness through pre-processing the graph data.
Due to the nature of pre-processing, our repairing procedure has no relationship with the embedding function $f$ and predictor $g$.
As a result, it becomes important to ensure that the repaired data can achieve dyadic fairness for arbitrary embedding function and predictor.
These can be considered as  the requirements \textbf{flexibility}.
Furthermore, another straightforward requirement needs to be emphasised, i.e., \textbf{unambiguity}. 
After repairing, the attribute and adjacency information of each node should be determined without ambiguity.

To obtain the wide applicability on predictors (flexibility), we consider optimizing the DBER of the most unfair predictor with the repaired data, i.e.,
\begin{equation}
\boldsymbol{Z}^* = \arg\max_{\boldsymbol{Z}} \min_{g}\  \mathrm{DBER}\left(g, \boldsymbol{Z}, \mathcal{S}\right).
\end{equation}
Suppose that the repaired data $\boldsymbol{Z}^*$ ensures high DBER under the most unfair predictor. In that case, it obtains dyadic fairness with wide applicability to predictors.
Although this makes the problem a bi-level optimization one, the closed form of $g$ can be obtained with the Bayes formula as in \cite{pmlr-v97-gordaliza19a}.
\begin{theorem}
The smallest DBER for the data $\boldsymbol{Z}$ is equal to:
\begin{equation}
\min_{g}\ \mathrm{DBER}\left(g, \boldsymbol{Z}, \mathcal{S}\right)=\frac{1}{2}\left(1-\frac{1}{2} W_{1.\neq}\left(\boldsymbol{\hat{\gamma}}_{0}, \boldsymbol{\hat{\gamma}}_{1}\right)\right),
\label{eq:rel}
\end{equation}
where $W_{1.\neq}$ denotes the Wasserstein distance between the conditional joint distributions of the node representation with the Hamming cost function.
$\boldsymbol{\hat{\gamma}}_0$ and $\boldsymbol{\hat{\gamma}}_1$ are conditional distributions over $\boldsymbol{Z}\times \boldsymbol{Z}$ given $S(u)\oplus S(v)=0$ and $S(u)\oplus S(v)=1$.
\end{theorem}
The detailed proof of this theorem has been elaborated in work \cite{pmlr-v97-gordaliza19a}.
As shown in the theorem, the dyadic balanced error rate of the most unfair predictor depends on the Wasserstein distance between the two conditional dyadic node representation distributions $(\boldsymbol{\hat{\gamma}}_0, \boldsymbol{\hat{\gamma}}_1)$.
When $W_{1.\neq}(\boldsymbol{\hat{\gamma}}_0,\boldsymbol{\hat{\gamma}}_1)=0$, which means that the two conditional distributions are identical, i.e.,
\begin{equation}
    \mathbb{P} \left( \boldsymbol{z}_u,\boldsymbol{z}_v\, \big|\, S(u)\oplus S(v)=1 \right) = \mathbb{P} \left( \boldsymbol{z}_u,\boldsymbol{z}_v\, \big|\, S(u)\oplus S(v)=0 \right).
    \label{eq: xor}
\end{equation}
The DBER can achieve the optimal $\frac{1}{2}$ and $\boldsymbol{Z}$ are taken as dyadic fairness on the sensitive feature $S$.
Ensuring \eqref{eq: xor} makes the repaired data achieve dyadic fairness with wide applicability on arbitrary predictor $g$.

\begin{figure}
\centering
\includegraphics[width=0.4\textwidth]{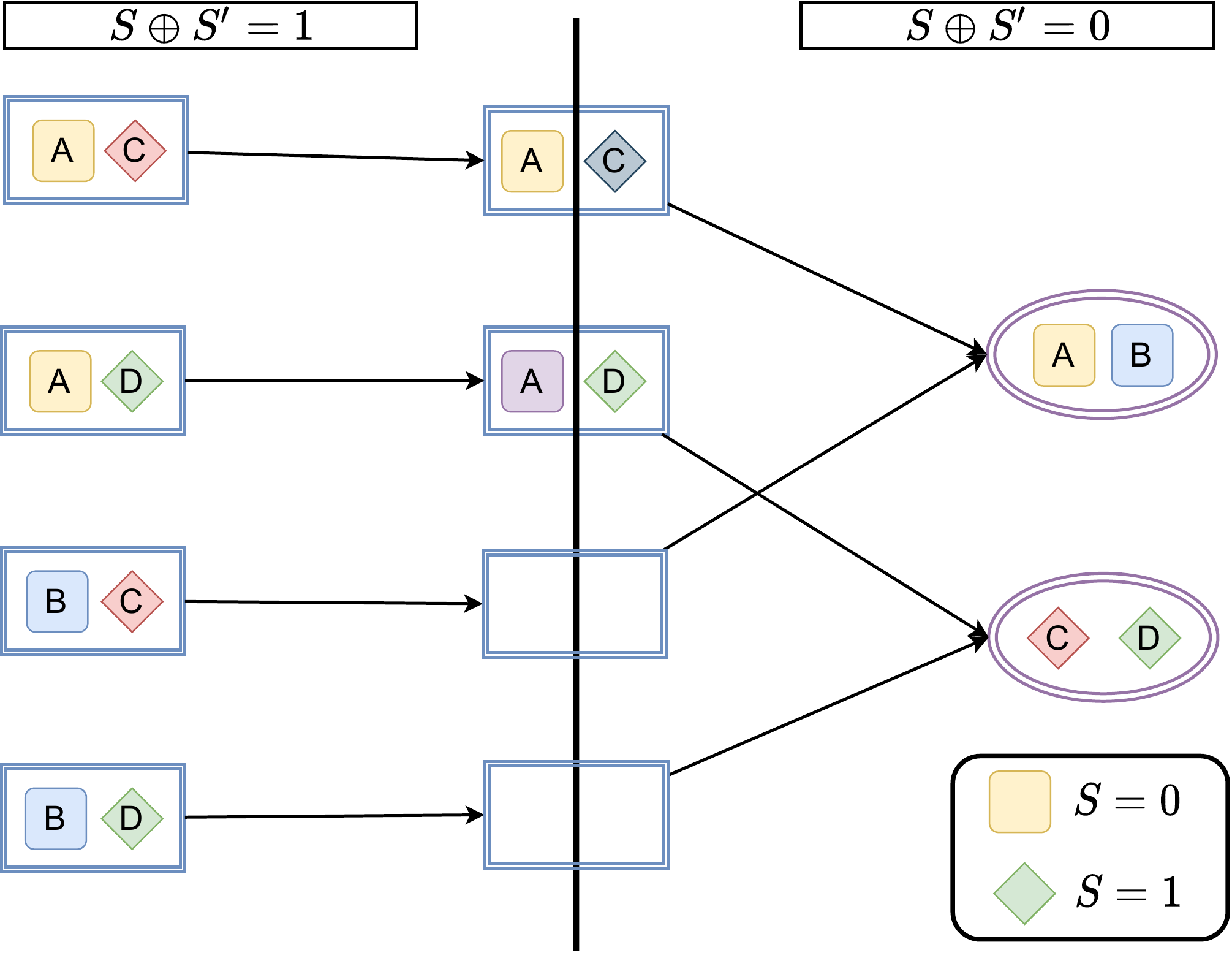}
\caption{The ambiguity illustration of dyadic repairing. These pairs $(A,C)$ and $(A,D)$ are repaired respectively to the pairs in the black line. $A$'s original attribute is yellow, while in the repaired data, it has multiple values (``yello'' and ``purple''), which might lead to ambiguity.}
\label{fig: illu}
\end{figure}
One straightforward repairing scheme is directly moving the two conditional distributions to the same distribution.
However, the representation of node $i$' $\boldsymbol{z}_i$ often occurs more times in $\boldsymbol{\hat{\gamma}}_0$ and $\boldsymbol{\hat{\gamma}}_1$.
When repairing $\boldsymbol{\hat{\gamma}}_0$ and $\boldsymbol{\hat{\gamma}}_1$, $\boldsymbol{z}_i$ will probably assign multiple values.
For example, as shown in Figure~\ref{fig: illu}, the direct repairing leads to ambiguity in the $A$'s attribute.
To achieve the unambiguity repairing, we propose the following proposition.
\begin{prop}
The dyadic fairness \eqref{eq: xor} is satisfied if and only if the following equation is satisfied:
\begin{equation}
    \mathbb{P} \left( \boldsymbol{z}_u\, \big|\, S(u)=0 \right) = \mathbb{P} \left( \boldsymbol{z}_v\, \big|\, S(v)=1 \right).
    \label{eq: emb}
\end{equation}
\end{prop}
\noindent{\bf{Proof}}: 
For the sufficient part, if \eqref{eq: emb} is satisfied, then for arbitrary representation $a$ and $b$, the
\begin{equation}
    \begin{aligned}
        &\ \mathbb{P} \left( \boldsymbol{z}_u = a, \boldsymbol{z}_v=b\, \big|\, S(u)\oplus S(v)=0 \right) \\
        = &{\textstyle{\sum_{i=0}^1}} \mathbb{P} \left( \boldsymbol{z}_u = a\, \big|\, S(u)=i \right) \times \mathbb{P} \left( \boldsymbol{z}_v=b \,\big| \, S(v)=i \right) \\
        = &{\textstyle{\sum_{i=0}^1}} \mathbb{P} \left( \boldsymbol{z}_u = a \,\big|\, S(u)=i \right) \times \mathbb{P} \left( \boldsymbol{z}_v=b \, \big| \, S(v)=1-i \right)\\
        = &\ \mathbb{P} \left( \boldsymbol{z}_u = a, \boldsymbol{z}_v=b \, \big| \, S(u)\oplus S(v)=1 \right),
    \end{aligned}
    \nonumber
\end{equation}which indicates the satisfactory of \eqref{eq: xor} accordingly.
For the necessary part, it can be easily proved by contradiction. 
The above proposition implies that a fair representation of nodes is sufficient to achieve dyadic fairness in the optimal case.
Repairing based on \eqref{eq: emb} allows us to obtain the dyadic fairness and unambiguity requirement due to the node's representation being only repaired once.
After achieving wide applicability on predictors and unambiguity, we consider obtaining the wide applicability on embedding function $f$.
The embedding function takes the whole graph $\mathcal{G}$ as input and outputs the node representation $\boldsymbol{z}_i$ based on the graph.
\begin{prop}
For any node $u$, $v$ in the graph $\mathcal{G}$, if they have the same node attributes and adjacency status, i.e,
\begin{equation}
    \mathbf{x}_u = \mathbf{x}_v \quad \mathrm{and} \quad \mathbf{e}_u = \mathbf{e}_v,
\end{equation}
then for any embedding function $f$, $f(\mathcal{G})[u] =f(\mathcal{G})[v]$.
$\mathbf{x}_u$, $\mathbf{x}_v$ denote the attribute of node $u$ and node $v$, respectively.
$\mathbf{e}_u$, $\mathbf{e}_v$ denote the 1-hop adjacency information, which means the local topology structure of node $u$ and node $v$.
\end{prop}
This proposition enables us to transform \eqref{eq: emb} into the following one: \begin{equation}
    \mathbb{P} \left( \mathbf{x}_u, \mathbf{e}_u \, \big| \, S(u)=0 \right) = \mathbb{P} \left( \mathbf{x}_v, \mathbf{e}_v \, \big| \, S(v)=1 \right).
    \label{eq: final}
\end{equation}
Based on \eqref{eq: final}, the dyadic fairness \eqref{eq: dp} can be further satisfied for arbitrary predictors.
In the following, we aim to propose an efficient algorithm to guarantee \eqref{eq: final}.

\section{Algorithmic Framework}

In this section, we introduce a practical and efficient algorithm called {\bf{DyadicOT}} to achieve dyadic fairness in link prediction tasks based on optimal transport theory.
It can be easily extended to multi-valued sensitive attributes problems, which can relax the binary sensitive value constraint.

\subsection{Dyadic fairness with optimal transport}
In order to achieve dyadic fairness through \eqref{eq: final}, we first represent the graph $\mathcal{G}$ as a matrix $\mathbb{R}^{N\times(d+N)}$ where each row represents the attribute of one node ($\mathbf{x}_u$) and adjacency information ($\mathbf{e}_u$).
According to the sensitive feature of each node, we further split $\mathcal{G}$ into $\mathcal{G}_0\in\mathbb{R}^{N_0\times (d+N)}$ and $\mathcal{G}_1\in\mathbb{R}^{N_1\times (d+N)}$ where $N_0$ and $N_1$ are the number of nodes with $S=0$ and $S=1$.
To bridge it with the optimal transport theory, we assume graph $\mathcal{G}_0$ and $\mathcal{G}_1$  form uniform distributions $\boldsymbol{\hat{\gamma}}_0$ and $\boldsymbol{\hat{\gamma}}_1$.
Our goal can be explicitly described as $\min_{\mathcal{G}}  W_{1.\neq}(\boldsymbol{\hat{\gamma}}_0, \boldsymbol{\hat{\gamma}}_1)$.
To achieve that goal, we solve the following optimal transport problem: 
\begin{equation}
    \boldsymbol{\Gamma}^* = \min_{\boldsymbol{\Gamma}\in\Pi \left( 1/N_0, 1/N_1 \right)} \left\langle \boldsymbol{\Gamma}, \mathbf{C} \right\rangle,
    \label{eq:opt}
\end{equation}
where $N_s$ is the number of nodes in the graph and $\frac{1}{N_s}$ is the uniform vector with $N_s$ elements, i.e., $s\in\{0,1\}$.

\subsubsection{Define the cost matrix ${\mathbf{C}}$}
Considering the distribution $\boldsymbol{\hat{\gamma}}_0$ and $\boldsymbol{\hat{\gamma}}_1$ encodes two important parts of information about the node, i.e., feature $\mathbf{x}_u$ and the local topology structure $\mathbf{e}_u$, our cost matrix $\mathbf{C}$ will consist of two components with hyperparameter $\eta$ as a trade-off between the feature term and the structure term.
\begin{equation}
    \mathbf{C}_{ij}=\eta \left\| \mathbf{x}_i,\mathbf{x}_j \right\|_2^2 + (1 - \eta) \left\|\mathbf{e}_i,\mathbf{e}_j\right\|_2^2.
    \label{eq:cost}
\end{equation}
To emphasis, although the Hamming distance is used in the above theoretical results, we practically employ the squared Euclidean distance.

\subsubsection{The DyadicOT algorithm}
The optimal transport plan $\boldsymbol{\Gamma}^*$ can be obtained, and further $\boldsymbol{\Gamma}^*$ can be utilized to repair the node feature and the adjacency information by mapping both $\mathcal{G}_0\in\mathbb{R}^{N_0\times (N+d)}$ and $\mathcal{G}_1\in\mathbb{R}^{N_1\times (N+d)}$ to the mid-point of the geodesic path between them \cite{villani2009optimal}, i.e.,
\begin{equation}
    \left\{\begin{array}{l}\tilde{{\mathcal{G}}_{0}}=\pi_{0} {\mathcal{G}}_{0}+\pi_{1} \boldsymbol{\Gamma}^{*} {\mathcal{G}}_{1}, \smallskip \\
    \tilde{{\mathcal{G}}_{1}}=\pi_{1} {\mathcal{G}}_{1}+\pi_{0} \boldsymbol{\Gamma}^{* \top} {\mathcal{G}}_{0}.\end{array}\right.
    \label{eq:repair}
\end{equation}
Following the above schemes \eqref{eq:opt}-\eqref{eq:repair}, the proposed DyadicOT algorithm can be concluded as follows.
\begin{algorithm}[ht]
\label{Algo1}
\caption{DyadicOT: Dyadic fairness with OT}
\begin{algorithmic}[1] 
\STATE Initialize $\eta$ and $\boldsymbol{\Gamma}^0\in\Pi\left( 1/N_0,1/N_1\right)$;
\STATE Split the graph $\mathcal{G}\in\mathbb{R}^{N\times (d+N)}$ into $\mathcal{G}_0\in\mathbb{R}^{N_0\times (d+N)}$ and $\mathcal{G}_1\in\mathbb{R}^{N_1\times (d+N)}$;
\STATE Compute the cost matrix $\mathbf{C}$ with \eqref{eq:cost};
\STATE Transform the distributions to their Wasserstein barycenter by solving \eqref{eq:opt};
\STATE Repair the $\mathcal{G}_0$ and $\mathcal{G}_1$ with \eqref{eq:repair}.
\end{algorithmic}
\end{algorithm}

\subsubsection{Multi-class extension}
In order to extend our approach to the case of the non-binary sensitive attribute, it would be necessary to compute the Wasserstein barycenter\cite{barycenter} of the conditional distributions.
Specifically, since each node has $|S|$ possible values of sensitive attribute, we first divide the graph $\mathcal{G}$ into $|S|$ sensitive attribute groups $\mathcal{G}_k\in\mathbb{R}^{N_k\times (d+N)}$ where $N_k$ is the number of nodes with $S=k$. 
Then, we compute the Wasserstein barycenter $\bar{\mathcal{G}}^*$ of these groups as follows:
\begin{equation}
    \bar{\mathcal{G}}^*=\underset{\bar{\mathcal{G}} \in \mathbb{R}^{N \times (N+d)}}{\operatorname{argmin}}\frac{1}{|S|}\sum_{k=1}^{|S|}\min _{\boldsymbol{\Gamma}_k \in \Pi\left(\frac{1}{N}, \frac{1}{N_{k}}\right)}\langle\boldsymbol{\Gamma}_k, \mathbf{C}_k\rangle,
\end{equation}
where $\mathbf{C}_k$ is the cost matrix between $\mathcal{G}_k$ and $\bar{\mathcal{G}}$. Once we have the Wasserstein barycenter $\bar{\mathcal{G}}^*$ and the optimal transport plan between the Wasserstein barycenter and each sensitive attribute group, i.e.,$\boldsymbol{\Gamma}_k$, we will repair each sensitive attribute group $\mathcal{G}_k$ as follows:
\begin{equation}
    \tilde{\mathcal{G}_k}=N_k{\boldsymbol{\Gamma}_{k}^{*}}^\top \bar{\mathcal{G}}^*.
\end{equation}

\section{Experiment Results}
This section specifies the experimental procedure of our approach on link prediction tasks and summarizes the analysis of the experimental results.

\subsection{Experiment Setup}
At the beginning, we first describe the experimental setup, including real-world datasets, baselines, evaluation metrics, and experiment details.

\smallskip
\noindent{\bf{Datasets}}. Our proposed algorithm is evaluated on two real-world network datasets. The statistical results for these two datasets are summarized in the following Table \ref{dataset}.

\begin{table}[ht]
\begin{center}
\caption{Statistic for datasets in experiments}
\label{dataset}
\begin{tabular}{@{}c|c|c|c|c@{}}
\midrule
Dataset  & \#Nodes & \#Edges & \#Node attributes & $|S|$   \\ \midrule
CORA     & $2708$   & $5278$   & $2879$   & $7$ \\ \midrule
CiteSeer & $2110$   & $3668$   & $3703$   & $6$ \\ \midrule
\end{tabular}
\end{center}
\end{table}



\noindent{-} {CORA}\footnote{{https://networkrepository.com/cora.php}} is a citation network consisting of $2708$ scientific publications classified into seven classes. Each node in the network is a publication, characterized by a bag-of-words representation of the abstract. The link between nodes represents undirected citations, and sensitive attributes are set to be the categories of the publication;

\noindent{-} {CiteSeer}\footnote{https://networkrepository.com/citeseer.php} dataset consists of $2110$ scientific publications classified into one of six classes. Similar to the CORA dataset, the node in the CiteSeer network is also a publication. Its sensitive attribute is set to be the publication's categories.


\smallskip
\noindent{\bf{Baselines}}. The following two pre-processing dyadic fairness baseline methods are chosen to be compared as follows:

\noindent{-} FairDrop\cite{DBLP:journals/corr/abs-2104-14210} is a biased dropout strategy that forces the graph topology to reduce the homophily of sensitive attributes.
Specifically, it generates a fairer random copy of the original adjacency matrix to reduce the number of connections between nodes sharing the same sensitive attributes;
\noindent{-} FairEdge\cite{DBLP:journals/corr/abs-2010-16326} is a theoretically sound embedding-agnostic method for group and individually fair edge prediction. 
It aims to repair the adjacency matrix of plain graphs based on the optimal transport theory and directly ignore the influence of node attributes.


\smallskip
\noindent{\bf{Evaluation metrics}}. In order to measure the structural changes between the repaired and the original graph for the pre-processing mechanism, we use Assortativity Coefficient (AC)\cite{DBLP:journals/corr/abs-2010-16326} to evaluate the correlation between the sensitive attributes of every pair of nodes that are connected. The values of AC always belongs to $[-1,1]$, and the value close to $0$ denotes that there is no strong association of the sensitive attributes between the connected nodes.

To evaluate the fairness, which is the main concern of our work, Representation Bias (RB)~\cite{buyl2020debayes} is employed to measure whether the embedding is well-obfuscated, i.e., contains no sensitive information.
Further more, we introduce a new dyadic fairness evaluation metric called DyadicRB through extending classical RB metric.
Similar with RB, the DyadicRB is calculated based on the accuracy of dyadic sensitive feature classification problem, which can be calculated as
$$
{\hbox{DyadicRB}} = \sum_{s=0}^{1}\frac{\big|\mathcal{E}_{s}\big|}{\big| \mathcal{E} \big|}\hbox{Accuracy} \left( S(u)\oplus S(v)\, \big|\, \boldsymbol{Z}_{u,v} \right).
$$
where $\boldsymbol{Z}_{u,v}$ is the edge embedding as the concatenation of the embeddings of the two nodes $u$ and $v$ connected by the link.
And $\hbox{Accuracy}(\cdot)$ is the accuracy of predicting the dissimilarity of sensitive information $S(u)\oplus S(v)$ based on edge embedding $\boldsymbol{Z}_{u,v}$.
Without limiting ourselves to unbiased embeddings, we utilize DDI \eqref{eq: di} to measure the fairness properties of the predictions themselves.
The effectiveness of our method on link prediction tasks from both the utility and fairness perspectives will be further evaluated.
As for the utility index, the Accuracy (ACC) is considered to measure the predictor's performance. 

\smallskip
\noindent{\bf{Experiment Details}}. Node2Vec\cite{grover2016node2vec} and support vector classifier are employed for all experiments as our embedding function and link predictor, respectively.
The dimension of the node`s embedding is $128$, and all values are collected with $5$ different random seeds.
For easy reproduction of the results, our codes are open-sourced in Github\footnote{https://github.com/mail-ecnu/OTDyadicFair}, and more details can be found there.

\subsection{Experiment Results}
In this section, we will evaluate and compare the effectiveness of our proposed DyadicOT method with other SOTA algorithms on real-world datasets at different stages along the pipeline of the link prediction task.

\smallskip
\noindent{\bf{Impact on the graph structure}}. Table \ref{ac} shows that the AC values of the two original graphs are relatively high, indicating that the links often appear between nodes with the same sensitive attributes.
This leads to discrimination against nodes with different sensitive attributes.
The three repairing methods can reduce the assortativity coefficient from the original graph.
Specifically, DyadicOT achieves smaller AC than FairEdge, which indicates the effectiveness of DyadicOT.
FairDrop could achieve a much smaller AC, and the resulting negative AC indicates that the different sensitive attribute nodes are more likely to connect.
However, the prediction accuracy of FairDrop may be highly influenced, and this phenomenon has been shown in Table \ref{cora} and Table \ref{citeseer}.

\begin{table}[tb!]
\centering
\caption{Assortativity Coefficient}
\label{ac}
\begin{tabular}{@{}c|c|c|c|c@{}}
\midrule
Dataset  & Original & FairEdge & FairDrop & {\bf{DyadicOT}} \\ \midrule
CORA     & $.771$   & $.668$   & $-.089$   & $.397$ \\ \midrule
CiteSeer & $.673$   & $.645$   & $-.065$   & $.567$ \\ \midrule
\end{tabular}
\end{table}

\smallskip
\noindent{\bf{Impact on node embeddings}}. Comparison on the impact on node embeddings among different repairing methods is another important concern.
Two aforementioned metrics are used, i.e.,
RB and DyadicRB, to quantify the fairness of the node embedding.
As shown in Tables~\ref{cora} and Table \ref{citeseer}, DyadicOT achieves the best score of both RB and DyadicRB.
These results indicate that both the sensitive attribute prediction and the dyadic sensitive attribute relation prediction are hard after repairing through DyadicOT.

\begin{table}[tb!]
\centering
\caption{Results on CORA. $\uparrow$ ($\downarrow$) denotes the higher (lower) the better respectively.}
\label{cora}
\scalebox{0.79}{
\begin{tabular}{@{}c|c|c|c|c@{}}
\midrule
             & ACC $\uparrow$& DDI$\uparrow$ & RB $\downarrow$& DyadicRB $\downarrow$  \\ \midrule
Original     & $\textbf{.829}\pm.007$ & $.266\pm.012$ & $.834\pm.004$ & $.726\pm.009$   \\ 
\midrule
FairEdge     & $.663\pm.008$ & $.393\pm.073$ & $.655\pm.004$ & $.596\pm.031$   \\ 
\midrule
FairDrop     & $.533\pm.019$ & $.657\pm.087$ & $.467\pm.015$ & $\textbf{.522}\pm.018$   \\ 
\midrule
DyadicOT         & $.614\pm.006$ & $\textbf{.836}\pm.106$ & $\textbf{.172}\pm.018$ & $\textbf{.522}\pm.013$ \\ \midrule
\end{tabular}
}
\end{table}

\begin{table}[tb!]
\centering
\caption{Results on CiteSeer.}
\label{citeseer}
\scalebox{0.79}{
\begin{tabular}{@{}c|c|c|c|cc@{}}
\midrule
             & ACC $\uparrow$& DDI$\uparrow$ & RB $\downarrow$& DyadicRB $\downarrow$&  \\ \cmidrule(r){1-5}
Original     & $.820\pm.011$ & $.372\pm.019$ & $.661\pm.005$ & $.658\pm.009$ &  \\ \cmidrule(r){1-5}
FairEdge     & $\textbf{.821}\pm.013$ & $.389\pm.018$ & $.655\pm.004$ & $.623\pm.023$ &  \\ \cmidrule(r){1-5}
FairDrop     & $.532\pm.024$ & $\textbf{.717}\pm.081$ & $.493\pm.021$ & $.510\pm.037$ &  \\ \cmidrule(r){1-5}
DyadicOT         & $.585\pm.014$ & $.653\pm.181$ & $\textbf{.211}\pm.027$ & $\textbf{.506}\pm.036$ &  \\ \cmidrule(r){1-5}
\end{tabular}
}
\end{table}

\begin{figure}[htbp]%
  \centering
  \subfloat[Original Embedding]{
        \includegraphics[width=0.2\textwidth]{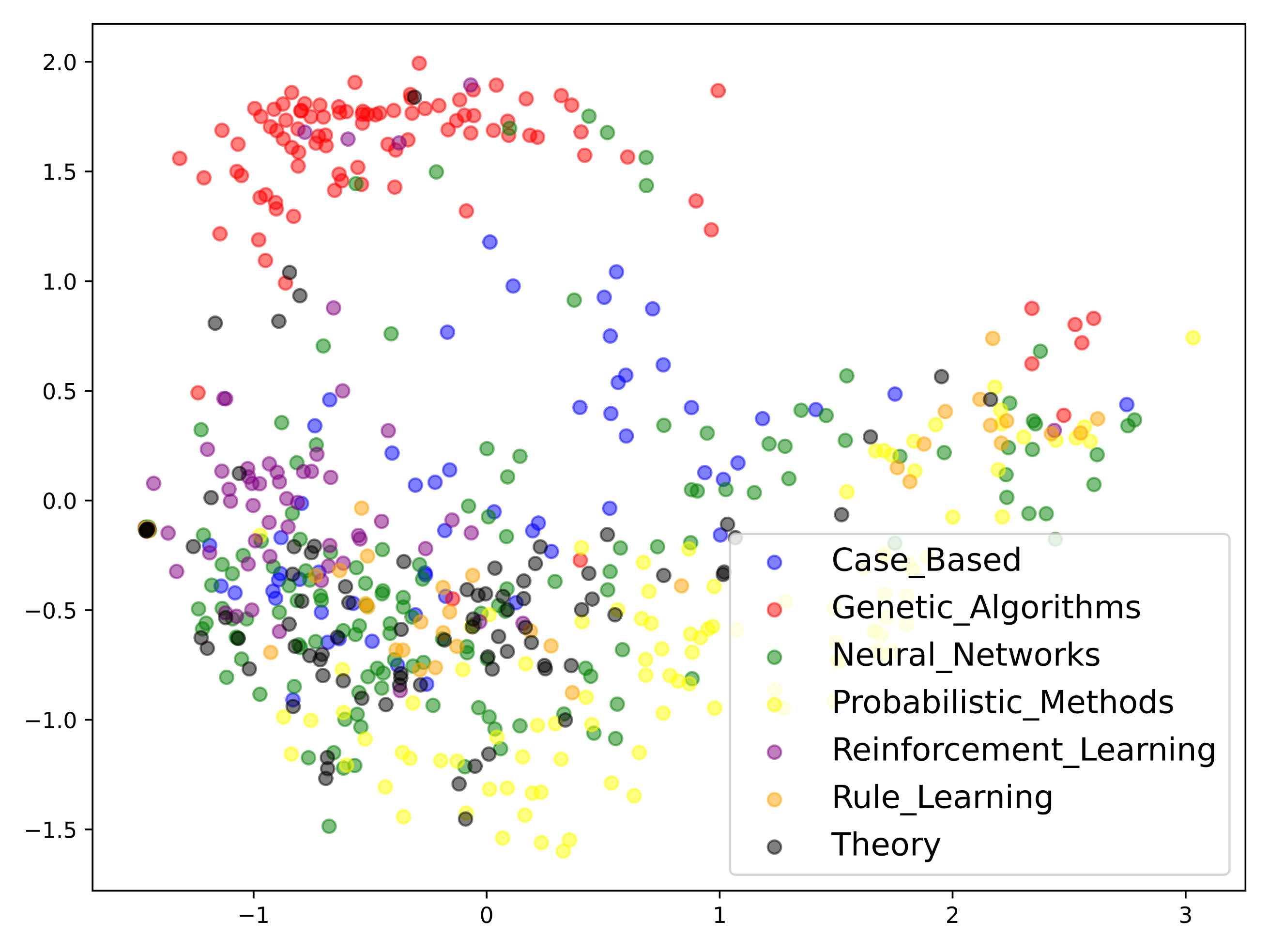}
        \label{fig:ori}
         }\hfill
  \subfloat[FairEdge's Embedding]{
        \includegraphics[width=0.2\textwidth]{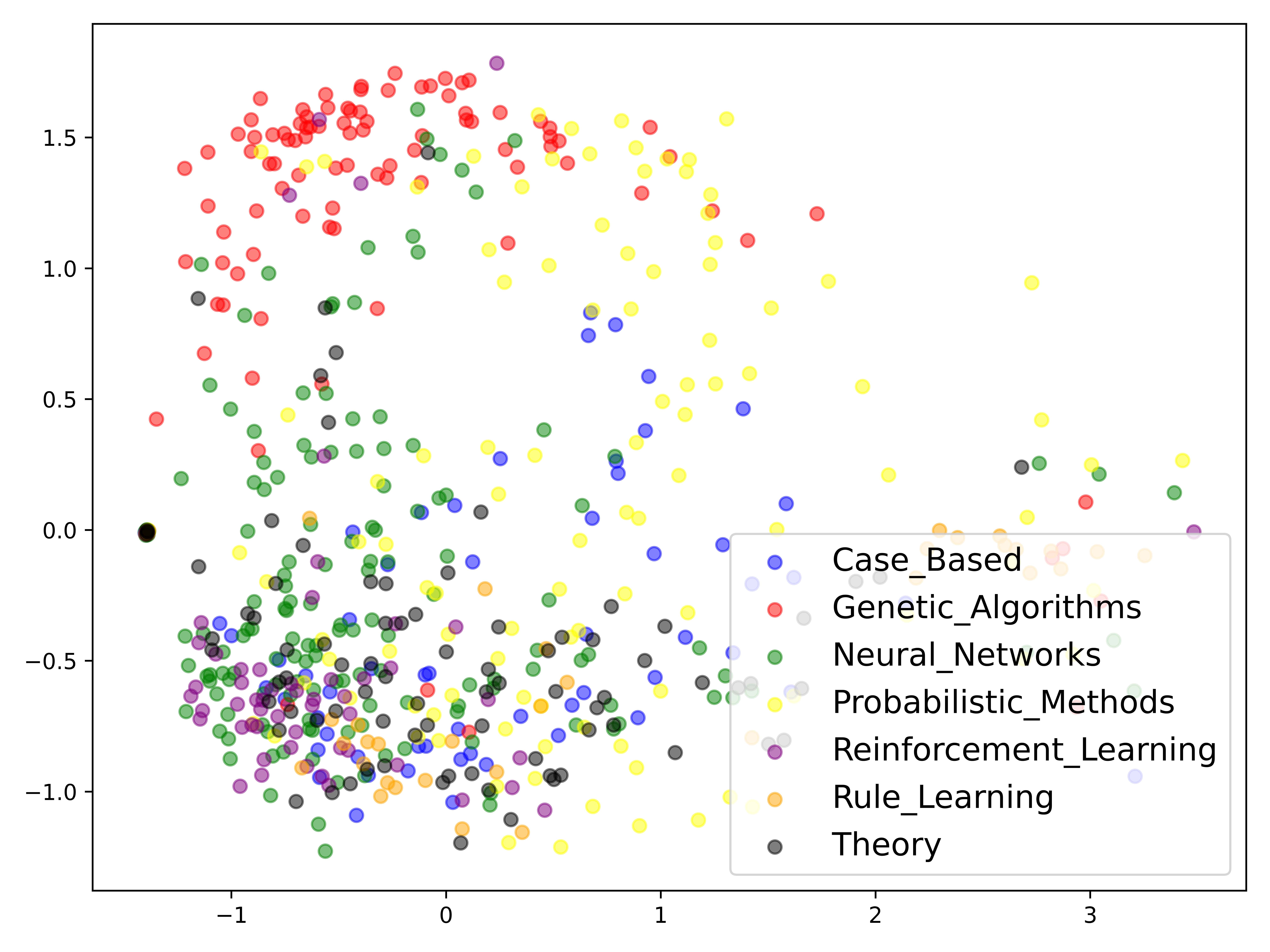}
        \label{fig:fairedge}
        }\\
  \subfloat[FairDrop's Embedding]{
        \includegraphics[width=0.2\textwidth]{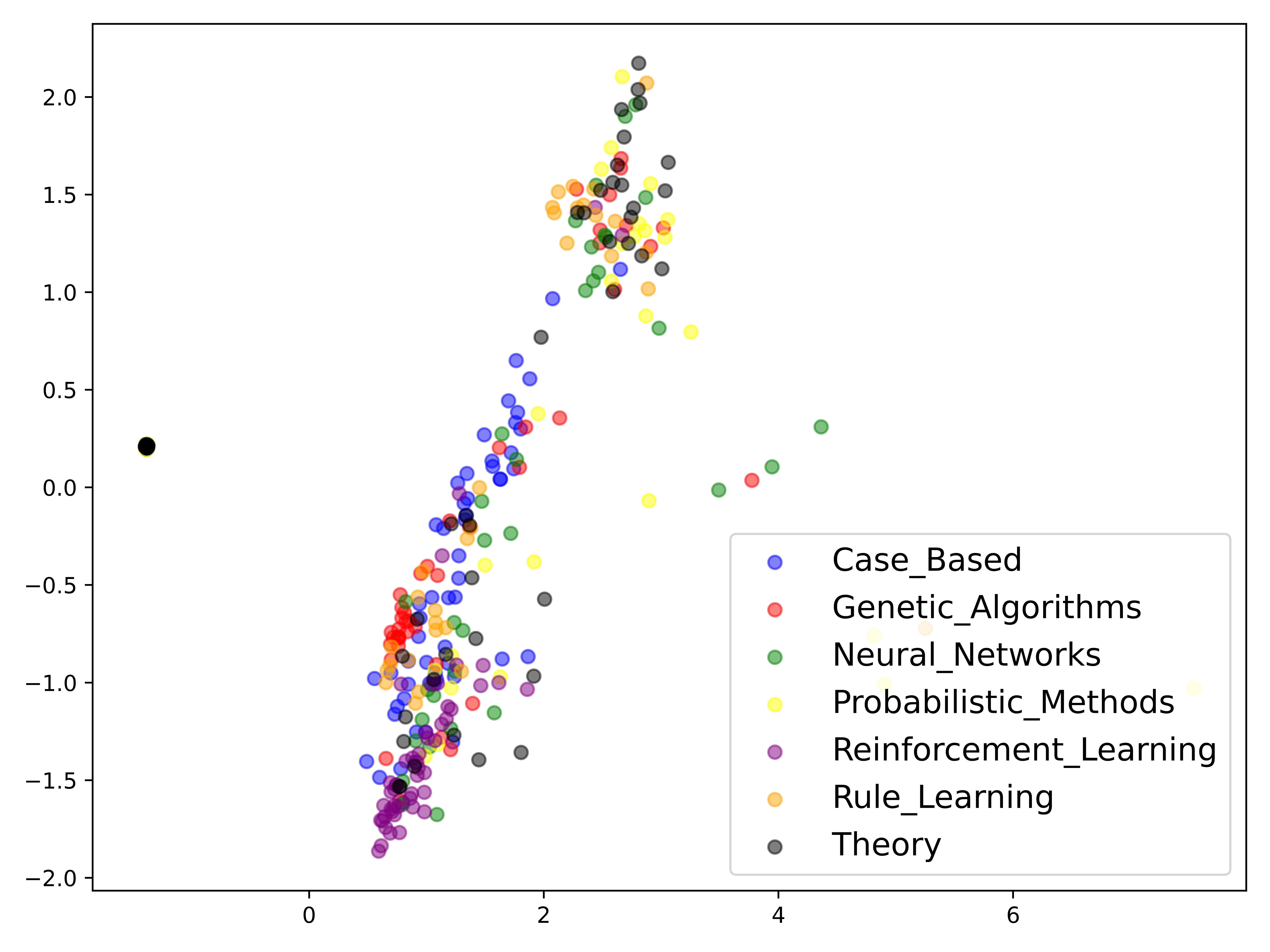}
        \label{fig:fairdrop}
         }\hfill
  \subfloat[DyadicOT's Embedding]{
        \includegraphics[width=0.2\textwidth]{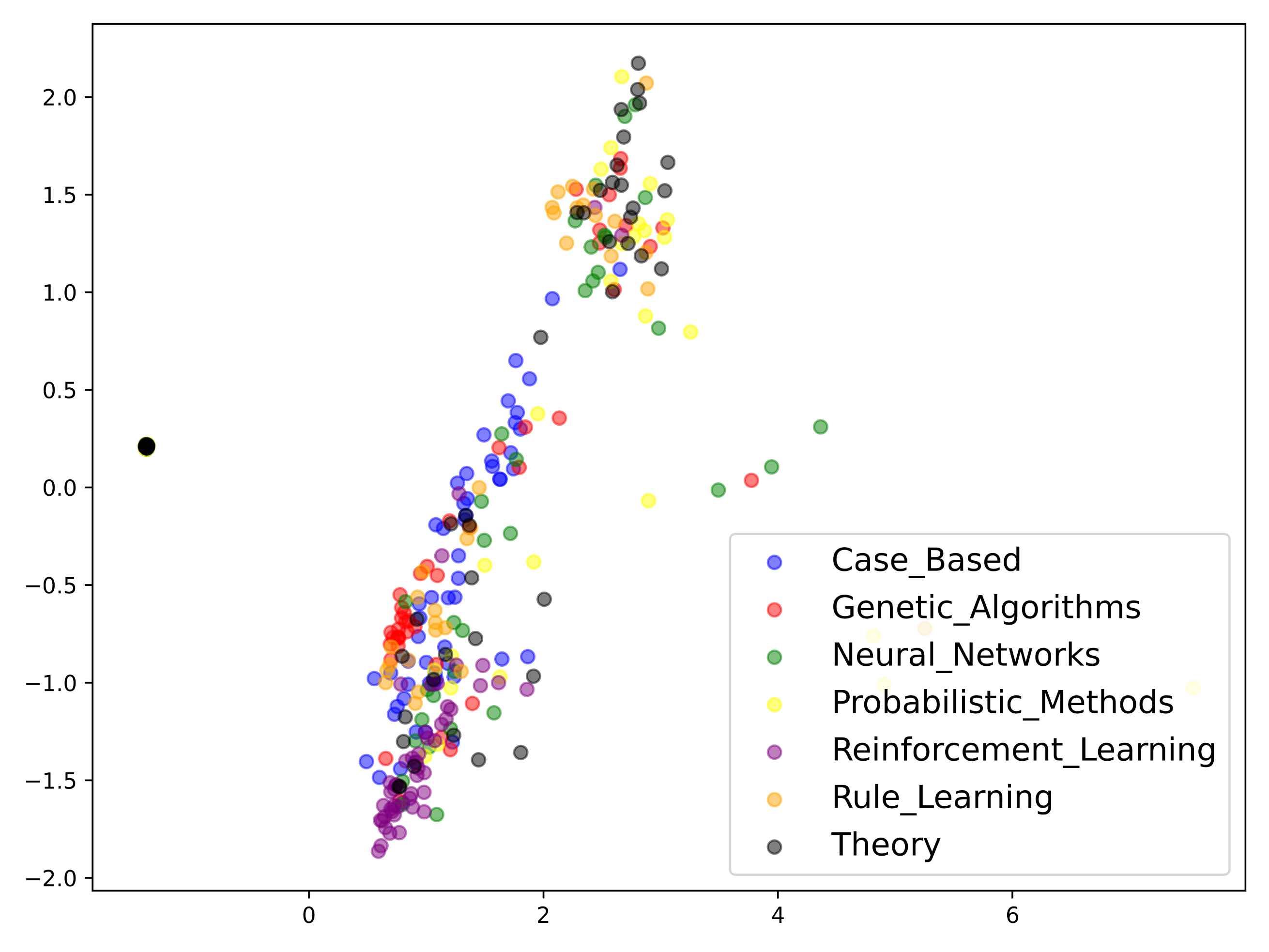}
        \label{fig:sym}
        }
  \caption{Visualization of node embedding learned by Node2Vec on CORA. Different colors indicate different sensitive attributes. (a) and (b) denote the node embeddings learned from the original graph or the graph repaired by DyadicOT respectively.}
\label{fig: sig}
\end{figure}

\begin{figure}[htbp]%
  \centering
  \subfloat[Original Embedding]{
        \includegraphics[width=0.2\textwidth]{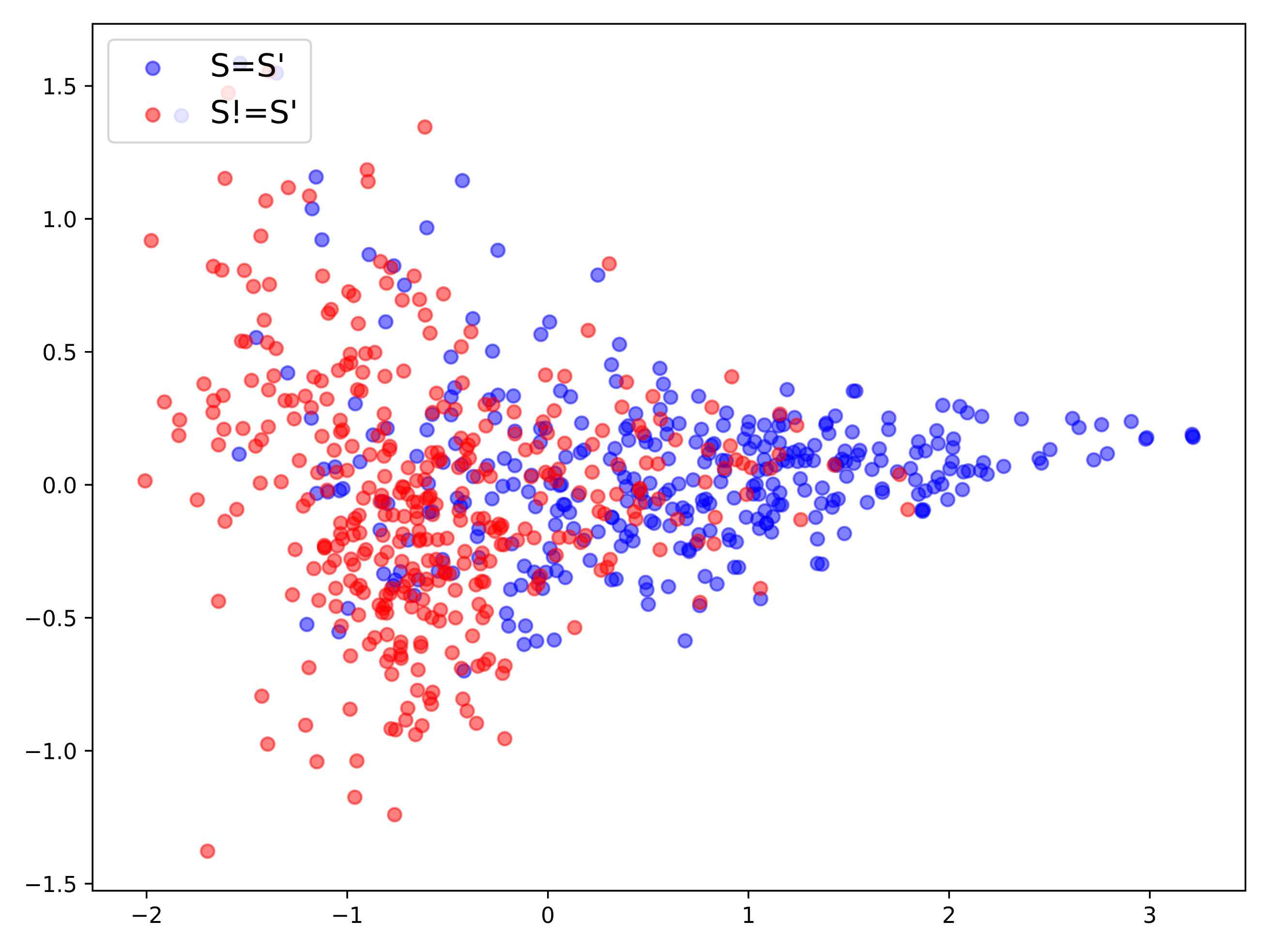}
        \label{fig:ori-pair}
         }\hfill
  \subfloat[FairEdge's Embedding]{
        \includegraphics[width=0.2\textwidth]{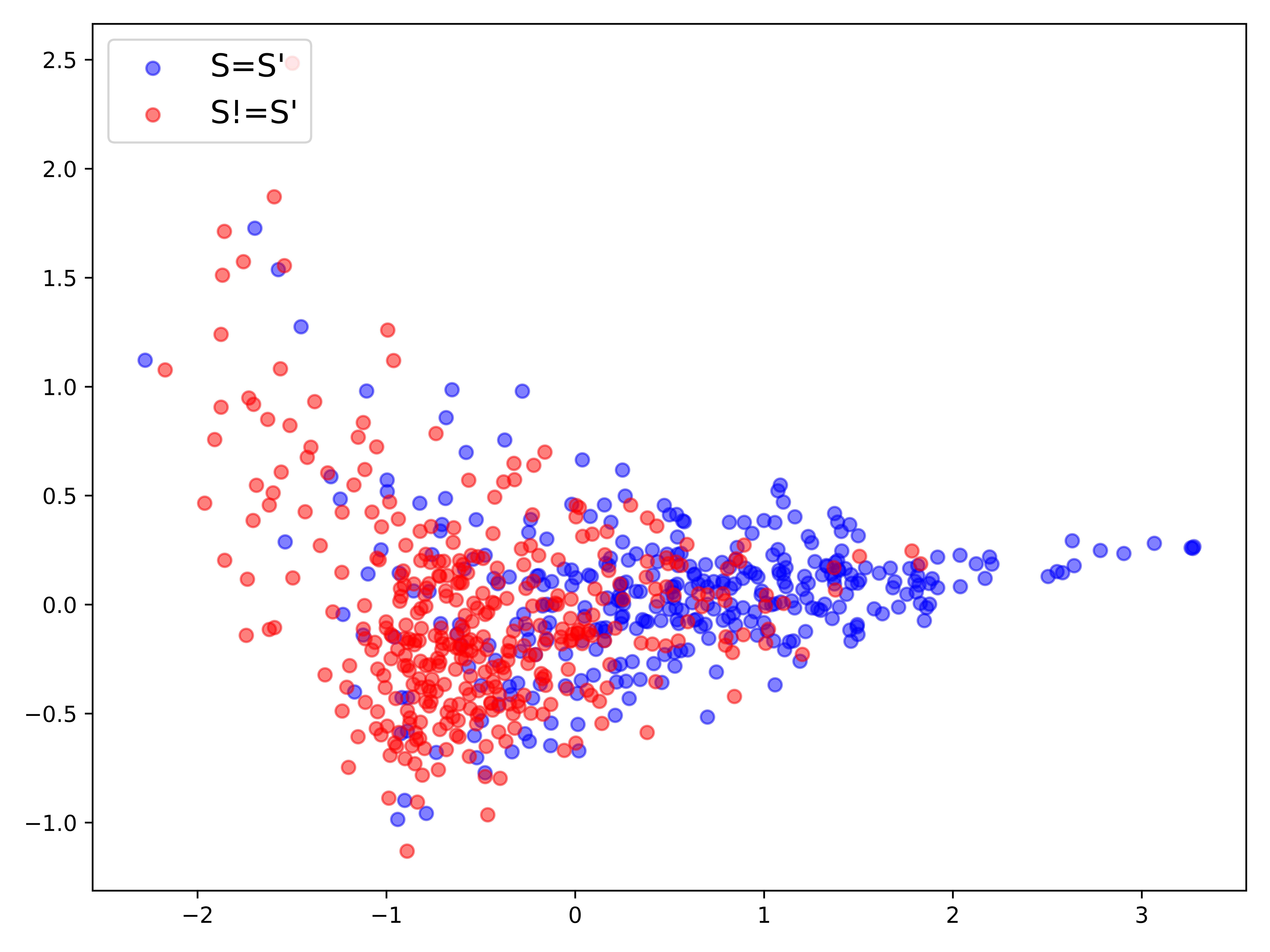}
        \label{fig:fairedge-pair}
        } \\
  \subfloat[FairDrop's Embedding]{
        \includegraphics[width=0.2\textwidth]{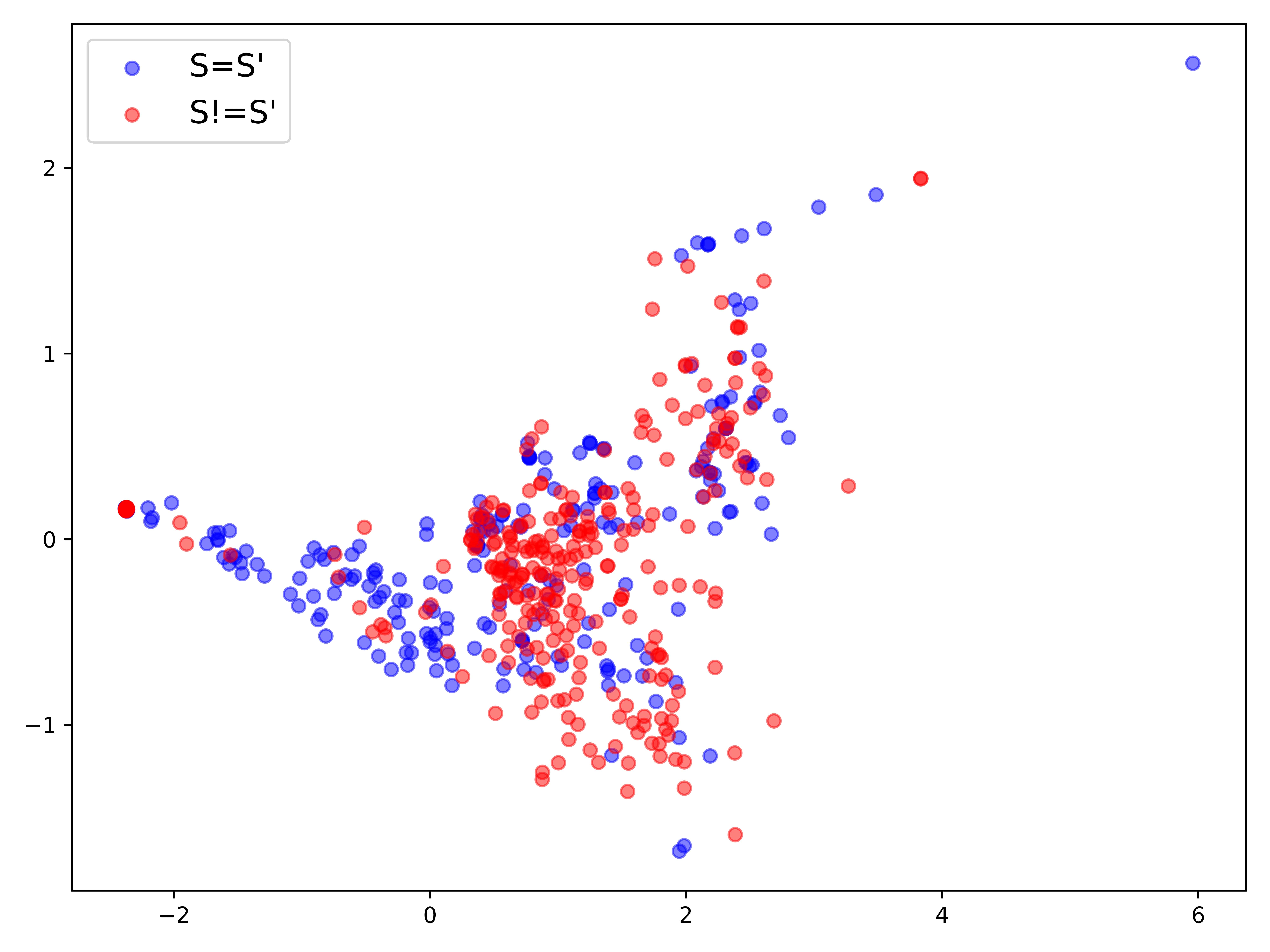}
        \label{fig:fairdrop-pair}
         }\hfill
  \subfloat[DyadicOT's Embedding]{
        \includegraphics[width=0.2\textwidth]{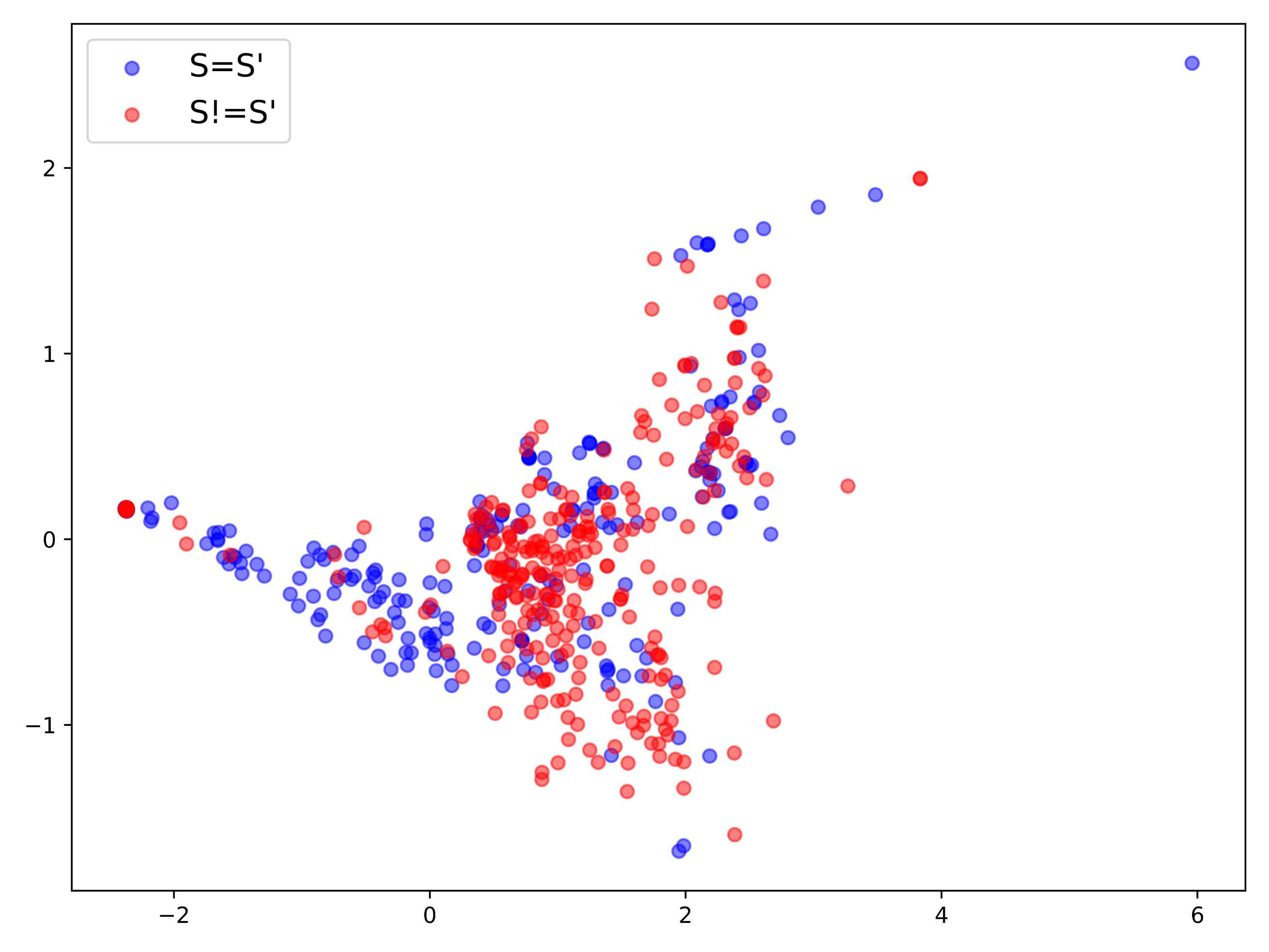}
        \label{fig:sym-pair}
        }
  \caption{Visualization of dyadic node embedding learned by Node2Vec on CORA. Here, the red colour represents node embeddings with different sensitive attributes, while the blue colour indicates node embeddings with the same sensitive attributes.}
  \label{fig: dyadic}
\end{figure}

To better understand the impact of our repairing on node embedding, we employ the PCA method to reduce the learned embedding into $2$-dimension space.
As shown in Figure~\ref{fig: sig}, the learned embedding from the original graph is distributed with highly correlated to the node's sensitive feature, which corresponds to higher RB.
The embedding learned from the repaired graph by DyadicOT is less correlated with the sensitive features compared with the baselines, corresponding to lower RB.
The comparison of dyadic embedding is shown in Figure~\ref{fig: dyadic}. 
The learned dyadic embedding by DyadicOT is less correlated than the original graph, indicating less predictability of the dyadic sensitive features' relationship (lower DyadicRB).

\smallskip
\noindent{\bf{Impact on link prediction}}. Finally, we consider the performance comparison on the link prediction task through two basic metrics, i.e., ACC and DDI.
ACC indicates the utility of the predictor, while DDI denotes the quantity of dyadic fairness the predictor achieves.
For the CORA dataset, all three repairing methods lose ACC while obtaining dyadic fairness.
Compared with FairEdge and FairDrop, DyadicOT achieves the best quantity of dyadic fairness (DDI), and the ACC decreases within the tolerance range.

As for the other CiteSeer dataset, FairEdge nearly cuts no ice on fairness.
However, 
compared to FairDrop, DyadicOT achieves higher DDI with less accuracy decrease, which indicates the better performance of DyadicOT.

\section{Conclusion}
This paper proposes a pre-processing method to achieve dyadic fairness in link prediction tasks.
By transforming the dyadic fairness obtaining problem into a conditional distribution alignment problem, dyadic fairness can be obtained with flexibility and unambiguity.
Furthermore, a practical repairing method is introduced based on optimal transport theory.
Experiments on CORA and CiteSeer show that the proposed DyadicOT method has significant results in obtaining the dyadic fairness of link prediction.




\end{document}